\documentclass{article}

\usepackage[final,nonatbib]{neurips_2019}

\usepackage[utf8]{inputenc}
\usepackage[T1]{fontenc}
\usepackage{url}
\usepackage{booktabs}
\usepackage{mathtools}
\usepackage{amsfonts}
\usepackage{nicefrac}
\usepackage[super]{nth}
\usepackage{hyperref}
\usepackage[capitalize, nameinlink, noabbrev]{cleveref}
\usepackage{microtype}
\usepackage{graphicx}
\usepackage{caption}
\usepackage{subcaption}
\usepackage{float}
\usepackage[section]{placeins}

\usepackage{fancyhdr}
\fancypagestyle{firstpage}{%
  \fancyhf{}

  \fancyhead[C]{\hspace{-1.3cm}Demo at NeurIPS 2019, Vancouver, BC, Dec. 2019.}
}

\title{Shared Mobile-Cloud Inference for Collaborative Intelligence}

\author{%
  Mateen Ulhaq and Ivan V. Baji\'{c} \\
  School of Engineering Science \\
  Simon Fraser University \\
  Burnaby, BC, V5A 1S6, Canada \\
  \texttt{mulhaq@sfu.ca, ibajic@ensc.sfu.ca} \\
}

\begin{document}

\maketitle

\thispagestyle{firstpage}

\begin{abstract}
  As AI applications for mobile devices become more prevalent, there is an
  increasing need for faster execution and lower energy consumption for neural
  model inference. Historically, the models run on mobile devices have been
  smaller and simpler in comparison to large state-of-the-art research models,
  which can only run on the cloud. However, cloud-only inference has drawbacks
  such as increased network bandwidth consumption and higher latency. In
  addition, cloud-only inference requires the input data (images, audio) to be
  fully transferred to the cloud, creating concerns about potential privacy
  breaches. We demonstrate an alternative approach: shared mobile-cloud
  inference. Partial inference is performed on the mobile in order to reduce
  the dimensionality of the input data and arrive at a compact feature tensor,
  which is a latent space representation of the input signal. The feature
  tensor is then transmitted to the server for further inference. This strategy
  can improve inference latency, energy consumption, and network bandwidth
  usage, as well as provide privacy protection, because the original signal
  never leaves the mobile. Further performance gain can be achieved by
  compressing the feature tensor before its transmission.
\end{abstract}

\section{Introduction}
\label{ssec:introduction}

\textit{Collaborative intelligence} is an AI model deployment strategy that
allows shared inference between the cloud and the  edge device. It has been
shown to provide benefits in terms of energy usage and inference latency in
certain scenarios~\cite{neurosurgeon}. Typically, an AI model, such as a deep
neural network, is split into an edge sub-model and a cloud sub-model. Feature
tensors computed by the edge sub-model are transmitted to the cloud for the
remainder of the inference process.

At NeurIPS 2019, we held a live demo to showcase an Android app demonstrating
shared inference strategies and comparing them against cloud-only and
mobile-only inference. ResNet~\cite{he2016deep} and VGG~\cite{simonyan2014deep}
models were preloaded onto the Android device and were run using the TensorFlow
Lite interpreter. We also had two servers emulating the cloud, each with a
GeForce GTX TITAN X GPU. One server was placed on our demo site's LAN and the
other was remote (located in a nearby city). Using the local server for
inference worked well for all configurations, but using the remote server for
inference only performed well for configurations that minimized the usage of
network bandwidth (e.g. shared inference strategies).

\begin{figure}[H]
  \centering
  \begin{subfigure}{.6\textwidth}
    \centering
    \includegraphics[width=.8\linewidth]{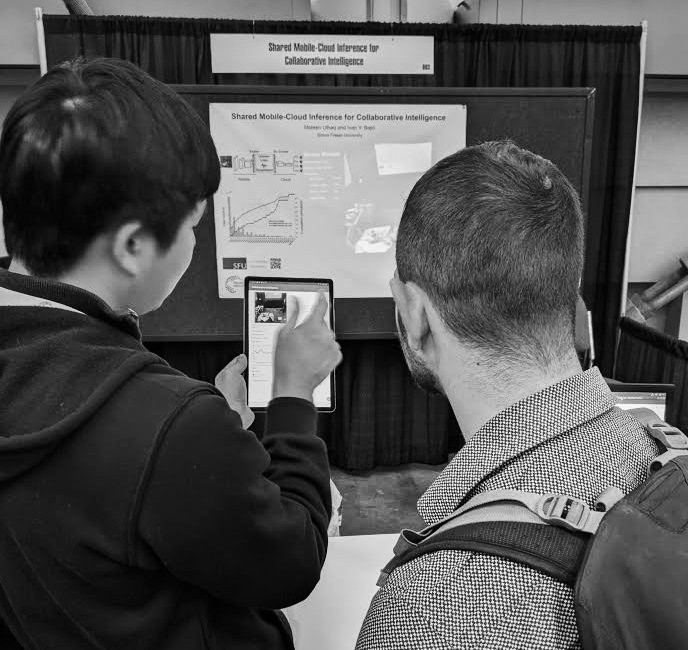}
    \caption[Live demo]{%
      Live demo at NeurIPS 2019.%
    }
    \label{fig:live_demo}
  \end{subfigure}%
  \begin{subfigure}{.4\textwidth}
    \centering
    \includegraphics[height=3in]{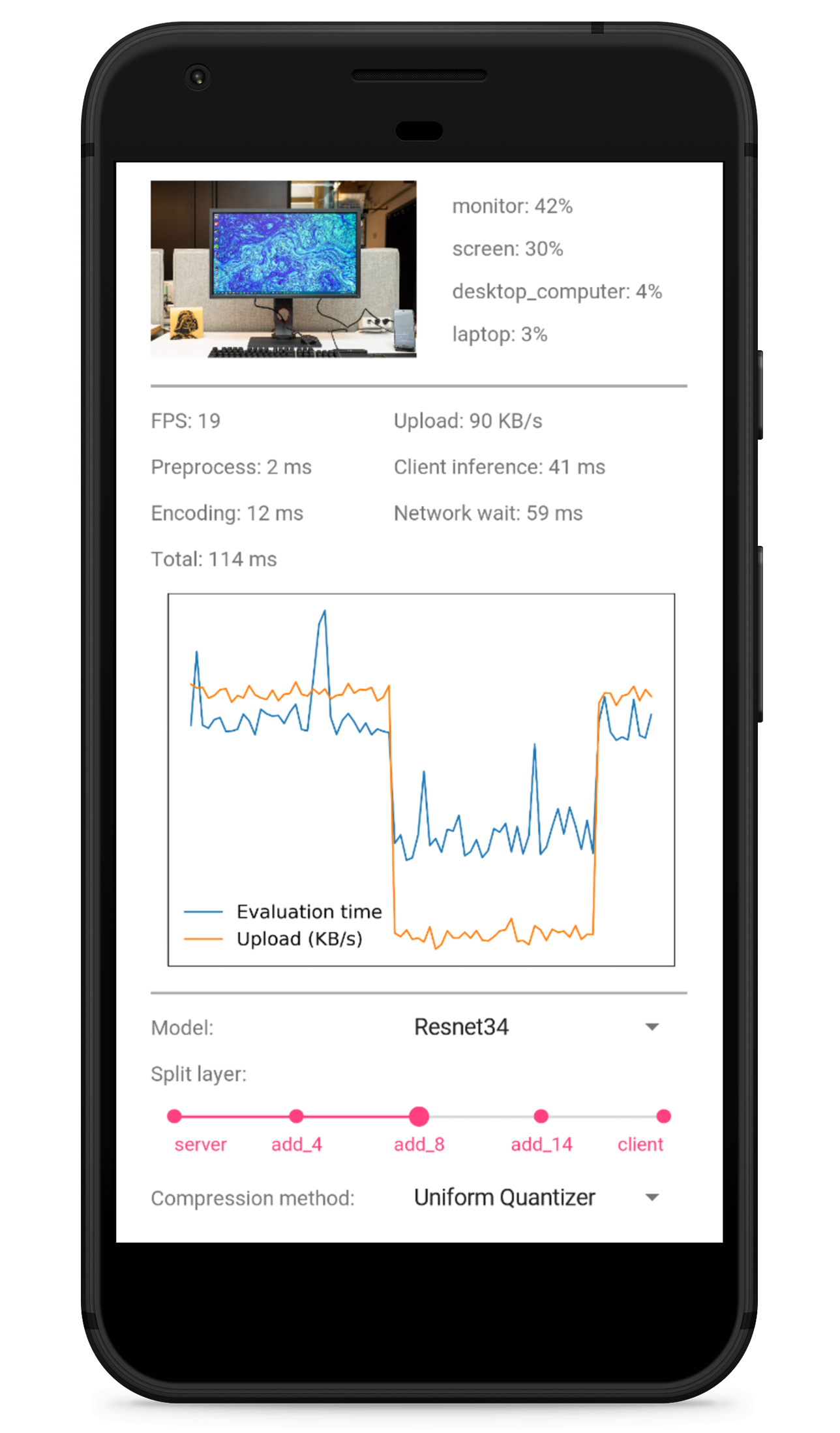}
    \caption[Android app]{%
      Demo app running on Android device. One may choose between the model,
      layer split points, and feature tensor compression method. Plots total inference time
      and tensor upload size for each frame.%
    }
    \label{fig:android_app}
  \end{subfigure}
\end{figure}

Inference on mobile GPUs tends to run slower than inference on server GPUs.
With our equipment, a single ResNet-34 inference took 160 ms on the mobile and
20 ms on the server. However, a server-only inference process is constrained by
the available bandwidth: low upload speeds can dramatically influence the total
end-to-end inference times. Furthermore, consumers also are wary of the
financial costs that come from mobile data usage. Shared inference strategies
attempt to lessen the negatives of server-only inference by transmitting less
data over the network, while still deferring heavier computations to the
server.

\begin{figure}
  \centering
  \includegraphics[width=\linewidth]{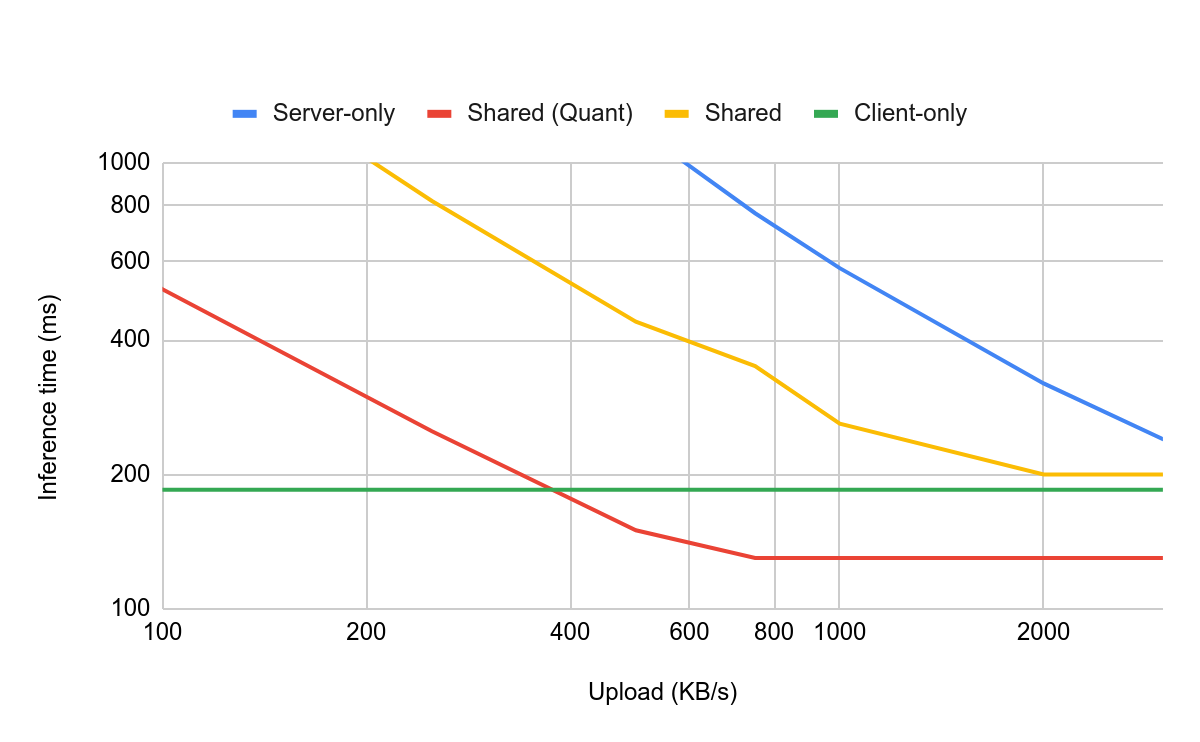}
  \caption[Measurements]{%
    A plot demonstrating how total inference time varies depending on available
    upload bit rate. Model under test: ResNet-34. Mobile client: Samsung Galaxy
    S10 phone (Qualcomm Snapdragon 845) on WiFi network with maximum available
    upload bit rate of 3 MB/s. Server: within 5 km and with average ping time
    of 5 ms.%
  }
  \label{fig:measurements}
\end{figure}

To demonstrate how shared inference may result in better overall inference
times, consider the plot in \cref{fig:measurements}.
For ResNet-34, shared inference using 8-bit quantized tensors is faster than
mobile client-only inference for upload bit rates larger than 450 KB/s, and
both are faster than cloud-only inference over the range of bit rates tested.
For this experiment, no further compression of tensors was applied, apart from
8-bit quantization.

These experiments were conducted on a Samsung Galaxy S10 phone running Qualcomm
Snapdragon 845 SoC, with maximum available upload bit rate of 3 MB/s; and a
remote server located within 5 km and average ping time of 5 ms. The inference
process for a new input image begins as soon as the previous image is
completely inferred. It is interesting to note that when using shared inference
strategies, it is possible to obtain higher throughputs by performing mobile
client-side inference on the next input image while waiting for the server to
respond.

A repository containing a demo Android app and Python library utilities to
assist in splitting models and conducting further analysis is made publicly
available.\footnote{%
\url{https://github.com/YodaEmbedding/collaborative-intelligence}}

\section{Methodology}
\label{ssec:methodology}

In order to deploy a deep neural network in a collaborative intelligence
setting, one needs to decide at which layer to split the model into the edge
sub-model and the cloud sub-model~\cite{neurosurgeon}. Desirable split layer
properties include:

\begin{itemize}
  \item Minimal aggregate computation from preceding layers
  \item Small output tensor size
  \item Compressibility of output tensor data
  \item Stability of tensor output values w.r.t. minor changes in input frames
\end{itemize}

\begin{figure}
  \centering
  \includegraphics[width=0.8\linewidth]{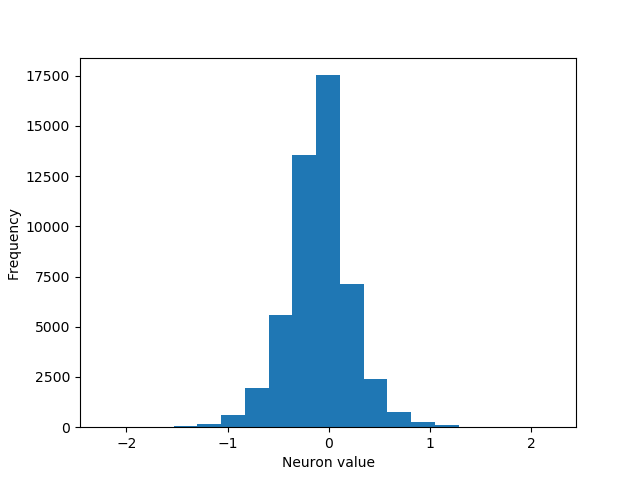}
  \caption[Layer output histogram]{%
    Histogram of output neuron values from \nth{8} add layer of ResNet-34.
    The neuron values resemble a normal distribution, and so nearly all neuron
    values are contained within the interval \protect$[-2, 2]$. This is largely
    due to the BatchNorm layer preceding the skip connection as well as the
    BatchNorm layer within the non-identity skip branch. These BatchNorm layers
    help ensure the output activations resemble a normal random variable.%
  }
  \label{fig:layer_histogram}
\end{figure}

Because the ResNet models contain BatchNorm \cite{DBLP:journals/corr/IoffeS15}
layers, the output neuron values of BatchNorm layers can be treated as a
normally distributed random variable, $y_{ijk} \sim \mathcal{N}(\mu = 0,
\sigma^2 = 1)$. Furthermore, the distribution of neuron values for the
following layers may also sometimes approximate a normal distribution. This is
visualized in \cref{fig:layer_histogram}. For a normal random variable, over
99\% of values lie within 3 standard deviations from the mean, so the tensor
can be quantized over the interval $[\mu - 3\sigma, \mu + 3\sigma]$. For the
demo app, we opted for 8-bit uniform quantization, though research
shows~\cite{DFTS} that even
fewer bits can be used without a significant drop in accuracy. Converting from
32-bit floating point values to 8-bit integers results in an immediate
4$\times$ reduction in the size of the tensor data. However, feature tensors
may be further compressed using standard codecs such as JPEG or H.264, or
codecs custom-build for deep feature tensor
data~\cite{choi_mmsp_2018,choi_icip_2018}. These can significantly improve
compression.
Further work is being done in our laboratory\footnote{SFU Multimedia Lab:
\url{http://multimedia.fas.sfu.ca/}} on developing custom codecs for deep
feature tensor compression. 

\section{Conclusion}

Using shared inference techniques, an edge device is able to perform
computationally expensive inferences under a larger variety of network
conditions than server-only inference is capable of, owing to the reduced
bandwidth usage. Moreover, the resulting inference times can be lower than
edge-only or cloud-only inference. Though we have only investigated shared
inference strategies with preexisting models such as ResNet, VGG, and
YOLO~\cite{redmon2018yolov3}, it is likely that models well-suited towards
shared inference strategies could be designed, containing layers that exhibit
the useful properties described in \cref{ssec:methodology}.

\bibliographystyle{ieeetr}
\bibliography{references}

\end{document}